\documentclass[lettersize,journal]{IEEEtran}
\usepackage{ifpdf}
\usepackage{cite}
\ifCLASSINFOpdf
  \usepackage[pdftex]{graphicx}
\else
  \usepackage[dvips]{graphicx}
\fi
\usepackage{amsmath}
\usepackage{algorithm}
\usepackage{amsthm}
\usepackage{amsfonts}
\usepackage{amssymb}
\usepackage{color}
\usepackage{algorithmic}
\usepackage{array}
\ifCLASSOPTIONcompsoc
\usepackage[caption=false,font=normalsize,labelfont=sf,textfont=sf]{subfig}
\else
    \usepackage[caption=false,font=footnotesize]{subfig}
\fi
\usepackage{fixltx2e}
\usepackage{stfloats}
\ifCLASSOPTIONcaptionsoff
    \usepackage[nomarkers]{endfloat}
    \let\MYoriglatexcaption\caption
    \renewcommand{\caption}[2][\relax]{\MYoriglatexcaption[#2]{#2}}
\fi
\usepackage{url}
\usepackage{algorithmic}
\begin{document}

%\begin{verbatim}
%\newtheorem{definition}{Definition} 
%\newtheorem{theorem}{Theorem}
%\end{verbatim}

\title{A Lyapunov Drift-Plus-Penalty Method Tailored for Reinforcement Learning with Queue Stability}

\author{Wenhan~Xu,~\IEEEmembership{Graduate Student Member,~IEEE,}
Jiashuo~Jiang,~\IEEEmembership{Member,~IEEE,}
Lei~Deng,~\IEEEmembership{Member,~IEEE,}
Yuan~Wu,~\IEEEmembership{Senior Member,~IEEE,}
and~Danny~H.K.~Tsang,~\IEEEmembership{Life Fellow,~IEEE}% <-this % stops a space
%\thanks{Manuscript received 29 May 2023; revised 29 September 2023; accepted 5 November 2023. This work was supported in part by, in part by, in part by, in part by, and in part by. \textit{(Corresponding author: Wenhan Xu.)}}

\thanks{Wenhan Xu is with the Internet of Things Thrust, Hong Kong University of Science and Technology (Guangzhou), Guangzhou, Guangdong 511400, China (e-mail: wxube@connect.ust.hk).}% <-this % stops a space
\thanks{Jiashuo Jiang is with Department of Industrial Engineering and Decision Analytics, Hong Kong University of Science and Technology, Hong Kong 99907, China (e-mail: jsjiang@ust.hk).}% <-this % stops a space
\thanks{Lei Deng is with the Internet of Things Thrust, Hong Kong University of Science and Technology (Guangzhou), Guangzhou, Guangdong 511400, China (e-mail: leideng@hkust-gz.edu.cn).}
\thanks{Yuan Wu is with The State Key Lab of Internet of Things for Smart City, and also with the Department of Electronic and Communication Engineering, University of Macau, Macao, China (e-mail: yuanwu@um.edu.mo).}
\thanks{Danny Hin-Kwok Tsang is with the Internet of Things Thrust, Hong Kong University of Science and Technology (Guangzhou), Guangzhou, Guangdong 511400, China, and also with the Department of Electronic and Computer Engineering, The Hong Kong University of Science and Technology, Clear Water Bay, Hong Kong SAR, China (e-mail: eetsang@ust.hk).}
%\thanks{Digital Object Identifier.}
%\thanks{Copyright (c) 2023 IEEE. Personal use of this material is permitted. However, permission to use this material for any other purposes must be obtained from the IEEE by sending a request to pubs-permissions@ieee.org.}
}

\markboth{}%
{Shell \MakeLowercase{\textit{et al.}}: Bare Demo of IEEEtran.cls for IEEE Journals}



\maketitle

\begin{abstract}
With the proliferation of Internet of Things (IoT) devices, the demand for addressing complex optimization challenges has intensified. The Lyapunov Drift-Plus-Penalty algorithm is a widely adopted approach for ensuring queue stability, and several researchers has preliminarily explored its integration with reinforcement learning (RL). In this paper, we investigate the adaptation of the Lyapunov Drift-Plus-Penalty algorithm for RL applications, deriving an effective method for combining Lyapunov Drift-Plus-Penalty with RL under a set of common and reasonable conditions through rigorous theoretical analysis. Unlike existing approaches that directly merge the two frameworks, our proposed algorithm, termed Lyapunov drift-plus-penalty method tailored for RL with queue stability (LDPTRLQ) algorithm, offers theoretical superiority by effectively balancing the greedy optimization of Lyapunov Drift-Plus-Penalty with the long-term perspective of RL. Simulation results for multiple problems demonstrate that LDPTRLQ outperforms the baseline methods using the Lyapunov drift-plus-penalty method and RL, corroborating the validity of our theoretical derivations. The results also demonstrate that our proposed LDPTRLQ outperforms other benchmarks in terms of compatibility and stability.

\end{abstract}


% \fi
\IEEEpeerreviewmaketitle

\section{Introduction}
The Lyapunov drift-plus-penalty framework has emerged as a widely adopted technique across diverse paradigms of communication systems, including backpressure routing \cite{nunez2012distributed}, Internet of Things (IoT) devices, adaptive video streaming \cite{li2018lyapunov}, and edge computing \cite{shuminoski2016lyapunov}. This method excels in stabilizing queues—both physical and virtual—within queueing networks and other stochastic systems while simultaneously optimizing time-averaged performance objectives, such as minimizing a penalty function \cite{yu2015convergence}. Its major strengths are implementation efficiency and simplicity. Unlike traditional approaches, the Lyapunov drift-plus-penalty method requires no prior knowledge of underlying statistical distributions (e.g., probability density functions of random channel conditions). Instead, it relies solely on greedy optimization of arbitrary utility functions at each time step. This lightweight and scalable design eliminates the need for complex probabilistic modeling, rendering the technique highly adaptable and broadly applicable \cite{hadi2019dynamic}.

Although the Lyapunov drift-plus-penalty method offers simplicity in implementation \cite{neely2022stochastic}, its effectiveness is often constrained by the complexity of penalty functions inherent in many optimization problems. These penalty functions, typically dictated by the optimization objectives of specific scenarios, are frequently complex and non-convex. While they remain computable, sometimes they do not inherently possess greedy properties, limiting the method’s applicability in such cases \cite{kerimov2025optimality}. Consequently, researchers have explored alternative approaches to address these challenges.

Recent advancements in data-driven reinforcement learning (RL) present a promising alternative for tackling online optimization problems in queueing systems \cite{yu2022deep}. This method autonomously learns from past experiences without requiring manually labeled training data, making it particularly well-suited for queueing problems that demand real-time, online implementation. Recent research has predominantly explored direct combinations of the Lyapunov drift-plus-penalty method and RL algorithms to tackle optimization challenges in stochastic systems \cite{huang2021multi}. The Lyapunov drift-plus-penalty method operates as a greedy algorithm, optimizing decisions based on immediate outcomes, whereas RL employs non-greedy exploration to maximize long-term rewards. As a result, devising a method that effectively unifies these two frameworks remains a complex challenge.

\subsection{Related Work}
The Lyapunov drift-plus-penalty method, first introduced by Neely in 2010 \cite{neely2010stochastic} to stabilize queues while optimizing specific objectives, has since found widespread application in queueing-related optimization problems, including backpressure routing \cite{nunez2012distributed}, IoT resource allocation \cite{kim2023lyapunov}, and edge computing \cite{xu2025lyapunov}. This approach transforms long-term queue stability goals into per-slot minimization of a drift-plus-penalty term, employing a greedy optimization strategy that requires no prior statistical knowledge of the system (e.g., stochastic probability density functions). Bracciale et al. \cite{bracciale2020lyapunov} developed a system based on this method, enforcing not only average queue stability but also strict per-slot maximum queue (or delay) capacity constraints. Similarly, \cite{jiang2022joint} proposed an online Lyapunov-based approach, investigating centralized and distributed schemes for joint task offloading and resource allocation in energy-constrained mobile edge computing (MEC) networks.
Furthermore, Hu et al. \cite{hu2024energy} implemented the Lyapunov drift-plus-penalty framework to obtain a dynamic resource allocation policy in a stochastic network optimization problem.

Concurrently, model-free RL algorithms, typically leveraging deep neural networks, learn effective policies through iterative trial-and-error interactions, rendering them well-suited for a variety of online optimization problems \cite{meng2024online}. In contrast to the Lyapunov drift-plus-penalty technique, which prioritizes per-slot optimization, RL focuses on maximizing long-term rewards, offering a fundamentally different approach to system dynamics. Implementing an RL framework requires only a standardized specification of actions, states, rewards, and state transitions \cite{wang2024offline}. The significance of RL in constrained markov decision processes (CMDP) contexts has garnered substantial attention in recent years, particularly for systems requiring safety or resource guarantees \cite{jiang2024achieving}. A variety of methods for tackling CMDP have been developed to balance objective maximization with constraint satisfaction\cite{efroni2020exploration,achiam2017constrained}. This versatility has led to its widespread adoption across diverse real-world applications, including large language models \cite{cao2024survey}, image recognition \cite{valente2023developments}, and perfect-information games \cite{nakayashiki2021maximum}.

Recent efforts have sought to integrate the Lyapunov drift-plus-penalty method with RL to harness their complementary strengths \cite{dai2020deep,bi2021lyapunov,liao2019learning,zhou2024predictable,jia2022lyapunov}. For instance, researchers in \cite{dai2020deep} incorporated Lyapunov optimization objectives directly into RL reward functions to address edge computing challenges in Digital Twin Networks. Similarly, Bi et al. \cite{bi2021lyapunov} applied a hybrid framework to mobile edge computing, using Lyapunov terms to guide RL policy updates. In \cite{liao2019learning}, a channel selection framework was proposed, combining machine learning, Lyapunov optimization, and matching theory to support industrial IoT implementations. 
Zhou et al. \cite{zhou2024predictable} proposed a predictable radio resource scheduling scheme leveraging Lyapunov-guided RL with proximal policy optimization. While the Lyapunov drift-plus-penalty framework provides a principled approach for long-term time-averaged optimization, its direct integration with RL poses practical challenges, where the greedy nature of the Lyapunov drift-plus-penalty method conflicts with RL’s focus on maximizing long-term rewards.

\subsection{Motivation and Contribution}
The rapid proliferation of portable IoT devices and applications has significantly increased the demand for optimization frameworks that ensure queue stability while addressing increasingly complex objectives. These scenarios, spanning domains such as backpressure routing and edge computing, often involve dynamic, non-convex optimization targets tied to practical performance metrics. Concurrently, advancements in artificial intelligence have prompted researchers to explore superior algorithmic alternatives, even in well-established fields like backpressure routing. The Lyapunov drift-plus-penalty method has emerged as a popular approach for optimizing application-specific objectives under queue stability constraints, leveraging its greedy, per-slot minimization strategy \cite{neely2010stochastic,bracciale2020lyapunov,jiang2022joint,dai2023uav,sun2024profit}. However, the dynamic, non-convex, and intricate nature of modern optimization goals has driven efforts to integrate this method with data-driven, model-free RL algorithms \cite{dai2020deep,bi2021lyapunov,liao2019learning,zhou2024predictable,jia2022lyapunov}. While such hybrid approaches have demonstrated performance gains in specific contexts, there exists a critical mismatch, namely, the greedy optimization inherent to the Lyapunov framework conflicts with RL's focus on maximizing long-term rewards, often leading to suboptimal or unstable outcomes. Our prior research investigated joint energy consumption optimization and queue stability in intelligent reflecting surface-assisted fog computing \cite{xu2024enhancing,xu2025reconfigurable}, highlighting the potential of tailored approaches in complex scenarios.

Motivated by the insights from the aforementioned literature review, we propose a novel reformulation of the Lyapunov drift-plus-penalty method, adapting it to align with RL's non-greedy paradigm. This paper presents several key contributions to address the integration of the Lyapunov drift-plus-penalty method with RL for queueing system optimizations.
\begin{itemize}
    \item  We conduct a theoretical analysis of the Lyapunov drift-plus-penalty method. Based on the practical significance of optimization objectives, we propose a set of basic theorems that such objectives should exhibit. Leveraging these theorems, we propose a Lyapunov drift-plus-penalty method tailored for RL with queue stability (LDPTRLQ) algorithm, ensuring compatibility with its non-greedy dynamics while preserving queue stability.
    \item We propose a dynamic Lyapunov weight selection algorithm to address scenarios where the Lyapunov weight is not predetermined in practical settings. Additionally, we introduce a saturated queue bailout mechanism that enables the LDPTRLQ algorithm to handle cases where the arrival rate of certain queues approaches the service rate limit.
    \item We validate the performance of the proposed RL-adapted Lyapunov drift-plus-penalty algorithm in a mobile edge computing problem, and compare it against the widely adopted original Lyapunov drift-plus-penalty-assisted RL approach. Numerical results demonstrate that our algorithm achieves better optimization performance and faster convergence. We also validate the performance of the proposed RL-adapted Lyapunov drift-plus-penalty algorithm in a routing problem. The results demonstrate that LDPTRLQ achieves performance comparable to the established backpressure routing algorithm while surpassing other hybrid approaches that integrate Lyapunov Drift-Plus-Penalty with RL. Across a variety of queueing optimization problems, our proposed LDPTRLQ algorithm demonstrates superior performance. These results underscore the algorithm's robust generality and scalability, highlighting its potential for broad applicability in diverse optimization scenarios.
\end{itemize}

\subsection{Notations}
Scalars and vectors are denoted by letters and bold letters, respectively. $|x|$ is the absolute value of the scalar $x$. $\Vert \boldsymbol{x}\Vert$ is the $L_2$-norm of vector $\boldsymbol{x}$. $\mathbf X^T$, $\mathbf X^H$, and $|\mathbf X|$ are the transpose, conjugate transpose, and determinant norm of a matrix $\mathbf X$, respectively. The primary notations in this paper are summarized in TABLE \ref{Parameters_notation}.

\section{System Model and Background}
In this section, we present the theoretical foundations of the Lyapunov drift-plus-penalty method applied to queueing systems. This framework provides a structured approach to stabilize queues while optimizing performance objectives, laying the groundwork for our subsequent analysis and algorithm design.

\begin{table}[t]
\renewcommand{\arraystretch}{1.3}
\caption{Parameters notation}
\label{Parameters_notation}
\centering
\begin{tabular}{cc}
\hline
\bfseries Parameters & \bfseries Notation\\
\hline
$n$ & $n^{th}$ queue\\
$N$ & Number of queues\\
$Q$ & Length of queue\\
$\mathbb{E}\{x\}$ & Expectation of $x$\\
$\mathbb{R}$ & Real number field\\
$d^{\text{arr}}$ & Arrival of the queue\\
$d^{\text{dep}}$ & Departure of the queue\\
$t$ & $t^{th}$ time slot\\
$\Delta T$ & Length of time slot\\
$\mathcal C$ & Lyapunov drift-plus-penalty function\\
$V$ & Weight between Lyapunov drift and penalty\\
$\overline p$ & Time-averaged Lyapunov penalty\\
$p(t)$ & Lyapunov penalty at time slot t\\%\mathcal
$\boldsymbol a$ & Action of RL algorithm\\
$\boldsymbol s$ & State of RL algorithm\\
$\mathcal A$ & Set of feasible actions\\
$\mathcal S$ & Set of feasible states\\
$\mathcal R$ & Reward of RL\\
$\lambda$ & Task arriving rate\\
$K$ & Number of mobile users\\

\hline
\end{tabular}
\end{table}

\subsection{Optimization Problem with Queueing Systems}
\begin{figure}[t]
\centering
\includegraphics[width=0.45\textwidth]{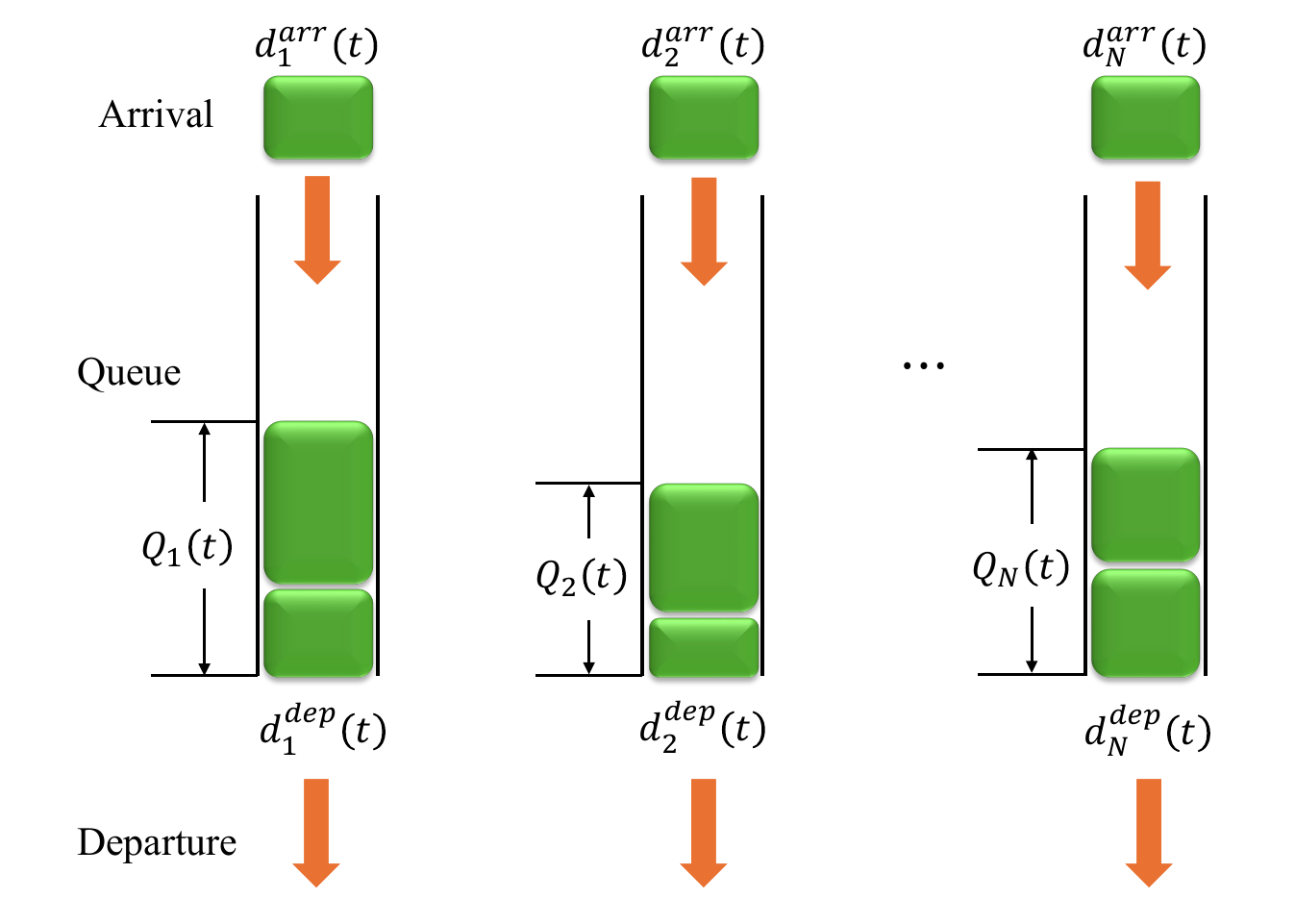}
\caption{System model of $N$ queues.}
\label{LDPP_f1}
\end{figure}
Consider a system comprising $N$ distinct queues. The queue lengths over time are represented by the following vector:
\begin{equation} 
	\boldsymbol Q(t)=[Q_1(t),Q_2(t),\cdots,Q_n(t),\cdots,Q_N(t)],
\end{equation}
where $Q_n(t)\geq0$ denotes the length of the $n^{th}$ queue at time slot $t\in\{0,1,2,3,\cdots\}$. To characterize the evolution of queue lengths, we denote the arrival process for the $n^{th}$ queue at time slot $t$ as $d^{\text{arr}}_n(t)$, and the departure process as $d^{\text{dep}}_n(t)$. The evolution of the queue lengths can be described by the following queueing model:
\begin{equation}
	Q_n(t+1)=\max\{Q_n(t)-d^{\text{dep}}_n(t),0\}+d^{\text{arr}}_n(t).
\end{equation}
\theoremstyle{definition}
\newtheorem{definition}{Definition}%[section]
\begin{definition}%[Fibration]
\cite{neely2022stochastic} A discrete time queue $Q_n$ is mean rate stable if
\begin{equation}
	\lim_{t\to\infty}\frac{\mathbb{E}\{Q_n(t)\}}{t}=0,
\end{equation}
\label{mean_rate_stable}
where $\mathbb{E}\{Q_n(t)\}$ is the expectation of $Q_n(t)$.
\end{definition}
After establishing the queue dynamics, we now formulate the optimization problem based on the Lyapunov drift-plus-penalty framework \cite{bracciale2020lyapunov}. The objective is to find a policy to minimize the average penalty with the mean rate stable queue:
\begin{equation}
    \begin{aligned}
		&\mathcal P1:\min\overline {p} \\
		\text {s.t.}\ &Q_{n} ~\text {is} ~\text {mean rate stable}, \forall n,\\
        &\alpha(t)\in\mathcal A(t),
    \end{aligned}
\end{equation}
where $\alpha(t)$ represents the action at time slot $t$ that the optimizer needs to decide, selected from the set of feasible actions $\mathcal A(t)$, and $\overline {p}$ is the time-averaged penalty excluding the queue stability constraint:
\begin{equation}
	\overline p=\lim_{\tau\to\infty}\frac{\sum_{t=0}^{\tau-1}\mathbb E\{p(t)\}}{\tau},
\end{equation}
where $p(t)$ denotes the penalty incurred at time slot $t$. Different specific problems have different $p(t)$ expressions. In practical applications, $p(t)$ typically represents the time-averaged cost-related metrics such as the energy consumption.

\subsection{Lyapunov Drift-Plus-Penalty Framework}
After establishing the basic optimization problem, we now turn to the Lyapunov drift-plus-penalty framework. The key here is the introduction of a Lyapunov function based on the queue vector $Q(t)$, defined as follows
\begin{equation}
	    LY(\boldsymbol Q(t))\triangleq \frac{1}{2}\sum_{n=1}^N Q^2_{n}(t),
\end{equation}
where $ LY(\boldsymbol Q(t))$ is the Lyapunov function of $Q(t)$. Accordingly, the Lyapunov drift at time slot $t$ is defined as the difference between the Lyapunov function values of consecutive time slots, expressed as
\begin{equation}
    \begin{aligned}
	    \Delta \boldsymbol Q(t)\triangleq& \mathbb E\{LY(\boldsymbol Q(t+1))-LY(\boldsymbol Q(t))|\boldsymbol Q(t)\}\\
        =&\mathbb E\{\frac{1}{2}\sum_{n=1}^N Q^2_{n}(t+1)-\frac{1}{2}\sum_{n=1}^N Q^2_{n}(t)|\boldsymbol Q(t)\}\\
        =&\mathbb E\{\frac{1}{2}\sum_{n=1}^N (Q^2_{n}(t+1)-Q^2_{n}(t))|\boldsymbol Q(t)\}.
    \end{aligned}
\end{equation}
Therefore, the Lyapunov drift-plus-penalty function at time slot $t$ is
\begin{equation}
    \begin{aligned}
	    \mathcal C(t)&=\underbrace{\Delta \boldsymbol Q(t)}_{\mathrm {Drift}}+\underbrace{V\mathbb E\{p(t)|\boldsymbol Q(t)\}}_{\mathrm {Penalty}}\\
        &=\underbrace{\mathbb E\{\frac{1}{2}\sum_{n=1}^N (Q^2_{n}(t+1)-Q^2_{n}(t))|\boldsymbol Q(t)\}}_{\mathrm {Drift}}\\
        &+\underbrace{V\mathbb E\{p(t)|\boldsymbol Q(t)\}}_{\mathrm {Penalty}},
    \end{aligned} 
\end{equation}
where $V$ is the weight between Lyapunov drift and penalty. By manipulating the controllable components of the optimization environment, the Lyapunov drift-plus-penalty method performs the following optimization problem (i.e., P2) at each time slot:
\begin{equation}
    \begin{aligned}
		&\mathcal P2:\min_{\alpha(t)}\mathcal C(t) \\
		&\text {s.t.}\ \alpha(t)\in\mathcal A(t),
    \end{aligned}
\end{equation}
where $\alpha(t)$ represents the action at time slot $t$ selected from the set of feasible actions $\mathcal A(t)$, and the Lyapunov drift-plus-penalty function $\mathcal C(t)$ is a function of the action $\alpha(t)$. Since the optimization process is confined to a single time slot and $Q(t)$ is a known quantity during per-slot optimization, the expectation terms in the Lyapunov drift-plus-penalty function $\mathcal C(t)$ can be replaced with their current values, yielding
\begin{equation}
    \begin{aligned}
		&\mathcal P3:\min_{\alpha(t)}\frac{1}{2}\sum_{n=1}^N (Q^2_{n}(t+1)-Q^2_{n}(t))+Vp(t) \\
		&\text {s.t.}\ \alpha(t)\in\mathcal A(t).
    \end{aligned}
\end{equation}
In several related works such as \cite{bracciale2020lyapunov,bi2021lyapunov}, the arrival length $d^{arr}(t)$ and departure length $d^{dep}(t)$ of queues within a single time slot are treated as bounded quantities and significantly smaller than the queue length $Q(t)$. Consequently, these existing works employ simplification and approximation techniques to transform problem $\mathcal P3$ into problem $\mathcal P4$ as follows
\begin{equation}
    \begin{aligned}
		&\mathcal P4:\min_{\alpha(t)}\sum_{n=1}^N Q_{n}(t)(d_n^{\text{arr}}(t)-d_n^{\text{dep}}(t))+Vp(t) \\
		&\text {s.t.}\ \alpha(t)\in\mathcal A(t).
    \end{aligned}
\end{equation}

\section{Lyapunov Drift-Plus-Penalty Method for RL}
Typically, the greedy optimization problems $\mathcal P3$ and $\mathcal P4$ can be solved directly. However, when $p(t)$ exhibits a complex, non-convex form, solving $\mathcal P3$ and $\mathcal P4$ may become NP-hard, rendering them computationally intractable. Furthermore, in practical optimization scenarios, the greedy minimization of the penalty $p(t)$ often fails to reduce its long-term time-averaged value, indicating weak greedy properties of the penalty term. In such cases, even if $\mathcal P3$ and $\mathcal P4$ are solvable, the resulting solutions yield suboptimal performance. This limitation has motivated researchers to explore model-free optimization techniques, such as deep learning. In this section, we aim to derive a theoretically sound adaptation of the Lyapunov drift-plus-penalty method for RL applications, addressing these challenges through rigorous analysis.
\subsection{Theoretical Derivation of the Proposed LDPTRLQ Method}
\subsubsection{Derivation of the form of optimization problem}
The following theorem builds on the widespread application of the Lyapunov drift-plus-penalty method to queue stability problems. This method's established efficacy in such contexts provides the foundation for our subsequent adaptation to RL frameworks.

\theoremstyle{theorem}
\newtheorem{theorem}{Theorem}%[section]
\begin{theorem}%[Fibration]
The Lyapunov drift-plus-penalty method provides an effective framework for addressing queue stability problems. Specifically, its effectiveness means that solving the optimization problem $\mathcal P3$ at each time slot can achieve the desired optimization objective while ensuring queue stability.
\label{LDPP_is_right}
\end{theorem}
Theorem \ref{LDPP_is_right} is proven in section 3.2 of \cite{neely2010stochastic}, and the successful applications of the Lyapunov drift-plus-penalty method in \cite{bracciale2020lyapunov,jiang2022joint,dai2023uav,sun2024profit} further validates Theorem \ref{LDPP_is_right}. In RL methodologies, the optimization objective is typically transformed directly into the reward function, with the action space and state space derived from the original optimization problem’s decision variables and feasible domain, respectively. A key advantage of RL is that it requires only the ability to compute the reward, which is equivalent to the optimization objective, rendering approximation or simplification of the reward unnecessary. Such simplifications, as applied in transforming $\mathcal P3$ to $\mathcal P4$, can even degrade RL efficiency. Consequently, we focus solely on the original problem $\mathcal P3$. Based on Theorem \ref{LDPP_is_right}, we deduce that the optimization problem we seek is equivalent to $\mathcal P3$. In terms of the equivalent optimization problem, we have Theorem \ref{theorem2}.
\begin{theorem}%[Fibration]
All problems equivalent to a given optimization problem can be characterized by applying a monotonically increasing function to its objective function:
\begin{equation}
\min_{\alpha\in\mathcal A} f(\alpha)\iff\min_{\alpha\in\mathcal A} \phi(f(\alpha))
\end{equation}
where $\phi$ is a monotonically increasing function with a single independent variable.
\label{theorem2}
\end{theorem}
\begin{proof}
We have an optimization problem $\mathcal{P}3a$ defined as
\begin{equation}
    \mathcal P3a:\min_{\alpha\in\mathcal A} f(\alpha),
\end{equation}
and another optimization problem $\mathcal{P}3b$ defined as
\begin{equation}
    \mathcal P3b:\min_{\alpha\in\mathcal A} g(\alpha).
\end{equation}
where $g(\alpha) = \phi(f(\alpha))$, and $\phi: \mathbb{R} \to \mathbb{R}$ is a monotonically increasing function. If the optimal solutions of $\mathcal P3a$ and $\mathcal P3b$ are identical, the equivalence can be proven. Let $\alpha^* \in \mathcal{A}$ be an optimal solution to $\mathcal{P}$, so that $f(\alpha^*) \leq f(\alpha), \forall \alpha \in \mathcal{A}$. Since $\phi$ is monotonically increasing, applying $\phi$ to both sides of the inequality preserves the order: $\phi(f(\alpha^*)) \leq \phi(f(\alpha)),\forall \alpha \in \mathcal{A}.$
Thus, there exists
\begin{equation}
g(\alpha^*) = \phi(f(\alpha^*)) \leq \phi(f(\alpha)) = g(\alpha), \quad \forall \alpha \in \mathcal{A}.
\end{equation}
This implies that $\alpha^*$ is also an optimal solution to $\mathcal{Q}$. Conversely, suppose $\beta^* \in \mathcal{X}$ is an optimal solution to $\mathcal{P}3b$, so that
\begin{equation}
g(\beta^*) = \phi(f(\beta^*)) \leq \phi(f(\alpha)) = g(\alpha), \quad \forall \alpha \in \mathcal{A}.
\end{equation}
Since $\phi$ is monotonically increasing, it is injective (one-to-one). Thus, $\phi(f(\beta^*)) \leq \phi(f(\alpha))$ implies $f(\beta^*) \leq f(\alpha)$ for all $\alpha \in \mathcal{A}$, because if $f(\beta^*) > f(\alpha)$, then $\phi(f(\beta^*)) > \phi(f(\alpha))$, contradicting the optimality of $\beta^*$ in $\mathcal{P}3b$. Hence, $\beta^*$ is also an optimal solution to $\mathcal{P}3a$. Therefore, the optimal solution sets of $\mathcal{P}3a$ and $\mathcal{P}3b$ are identical, proving that $\mathcal{P}3b$ is equivalent to $\mathcal{P}3a$.

\end{proof}
Indeed, all optimization problems equivalent to $\mathcal P3$ with different objectives can be expressed in the following form
\begin{equation}
    \begin{aligned}
		&\mathcal P5:\min_{\alpha(t)}f_{1,t}(\frac{1}{2}\sum_{n=1}^N (Q^2_{n}(t+1)-Q^2_{n}(t))+Vp(t)) \\
		&\text {s.t.}\ \alpha(t)\in\mathcal A(t),
    \end{aligned}
\end{equation}
where $f_{1,t}, \forall t$ is a monotonically increasing function with a single independent variable. Since $f_{1,t}(x), \forall t$ achieves its minimum value when $x$ attains the minimum within its domain, this transformation preserves the solution to the minimization problem.
\subsubsection{Derivation of undetermined term in the optimization problem}

The reward function for the RL algorithm we seek for $\mathcal P5$ can be formulated as follows
\begin{equation}
	    \mathcal R(t)=-f_{1,t}(\frac{1}{2}\sum_{n=1}^N (Q^2_{n}(t+1)-Q^2_{n}(t))+Vp(t)).
        \label{ALL_R}
\end{equation}
\theoremstyle{lemma}
\newtheorem{lemma}{Lemma}
According to Theorem \ref{theorem2} and its proof, we have Lemma \ref{lemma2}.
\begin{lemma}%[Fibration]
Eq. (\ref{ALL_R}) represents the set of all reward functions that satisfy the conditions of Theorem \ref{LDPP_is_right}.
\label{lemma2}
\end{lemma}

We will gradually narrow down the scope of the set in Lemma \ref{lemma2} based on some properties that RL should satisfy when dealing with this problem. We also establish the following Theorem \ref{theorem3}.
\begin{theorem}%[Fibration]
Existence theorem: Since Eq. (\ref{ALL_R}) represents the set of all reward functions, there exists at least one form of Eq. (\ref{ALL_R}) that is an optimal and static reward function.
\label{theorem3}
\end{theorem}
The premise for Theorem \ref{theorem3} to hold is that the Lyapunov drift plus penalty algorithm can be used in RL. Many studies, e.g., \cite{bracciale2020lyapunov,jiang2022joint,dai2023uav,sun2024profit}, have confirmed this. Therefore, we regard Theorem \ref{theorem3} as an axiom. As our goal is to devise a general method applicable to arbitrary environments, we can first focus on a special case where the queue length is independent of the action $\alpha$. By Theorem \ref{LDPP_is_right}, in this case, the greedy policy that minimizes the per-slot objective in each time slot is optimal with respect to the long-run time-average objective.
We have the original problem $\mathcal P6$ and the target problem $\mathcal P7$ suitable for RL:
\begin{equation}
    \begin{aligned}
		&\mathcal P6:\min_{\alpha(t)}Vp(t)+C(t) \\
		&\text {s.t.}\ \alpha(t)\in\mathcal A(t),
    \end{aligned}
\end{equation}
and
\begin{equation}
    \begin{aligned}
		&\mathcal P7:\min_{\alpha(t)}f_{1,t}(Vp(t)+C(t)) \\
		&\text {s.t.}\ \alpha(t)\in\mathcal A(t),
    \end{aligned}
\end{equation}
where $C(t)=\frac{1}{2}\sum_{n=1}^N (Q^2_{n}(t+1)-Q^2_{n}(t))$ is independent of the action $\alpha$. Since $C(t)$ is a constant independent of the optimization variables, and by Theorem \ref{theorem2}, it can be safely removed from the objective function without affecting the optimal solution. Consequently, the term $C(t)$ is omitted, yielding the simplified problems $\mathcal P6a$ and $\mathcal P7a$ as follows:
\begin{equation}
    \begin{aligned}
		&\mathcal P6a:\min_{\alpha(t)}Vp(t) \\
		&\text {s.t.}\ \alpha(t)\in\mathcal A(t),
    \end{aligned}
\end{equation}
and
\begin{equation}
    \begin{aligned}
		&\mathcal P7a:\min_{\alpha(t)}f_{1,t}(Vp(t)) \\
		&\text {s.t.}\ \alpha(t)\in\mathcal A(t).
    \end{aligned}
\end{equation}

The greedy minimization of the original objective achieves the minimization of its long-term time-averaged value. To achieve the long-term time-averaged value of $\mathcal P6a$, we need to solve
\begin{equation}
    \begin{aligned}
		&\mathcal P8:\min_{\alpha}\lim_{\tau\to\infty}\frac{\sum_{t=0}^{\tau-1}Vp(t)}{\tau} \\
		&\text {s.t.}\ \alpha\in\mathcal A.
    \end{aligned}
\end{equation}
According to Theorem \ref{theorem2}, $\mathcal P8$ is equivalent to
\begin{equation}
    \begin{aligned}
		&\mathcal P9:\min_{\alpha}\lim_{\tau\to\infty}\frac{f_{1,t'}(\sum_{t=0}^{\tau-1}Vp(t))}{\tau},\forall t', \\
		&\text {s.t.}\ \alpha\in\mathcal A,
    \end{aligned}
\end{equation}
where $t'$ can be any integer. The RL algorithm to solve $\mathcal P7a$ will look for the minimum long-term time-averaged value
\begin{equation}
    \begin{aligned}
		&\mathcal P10:\min_{\alpha}\lim_{\tau\to\infty}\frac{\sum_{t=0}^{\tau-1}f_{1,t}(Vp(t))}{\tau} \\
		&\text {s.t.}\ \alpha\in\mathcal A.
    \end{aligned}
\end{equation}
According to Theorem \ref{theorem2}, $\mathcal P10$ is equivalent to
\begin{equation}
    \begin{aligned}
		&\mathcal P11:\min_{\alpha}\lim_{\tau\to\infty}\frac{f_2(\sum_{t=0}^{\tau-1}f_{1,t}(Vp(t)))}{\tau} \\
		&\text {s.t.}\ \alpha\in\mathcal A,
    \end{aligned}
\end{equation}
where $f_2$ is a monotonically increasing function with a single independent variable. According to the assumption of this special case when the queue length is independent of the action $\alpha$, $\mathcal P11$ is equivalent to $\mathcal P9$. Therefore,
\begin{equation}
	    f_{1,t'}(\sum_{t=0}^{\tau-1}Vp(t))=f_2(\sum_{t=0}^{\tau-1}f_{1,t}(Vp(t))),\forall t'.
        \label{p9_p11}
\end{equation}
Eq. (\ref{p9_p11}) holds true for any value of $\{p(t),\tau,V\}$.
Let $\tau=1, t'=0$, we can get
\begin{equation}
	    f_{1,0}(Vp(0))=f_2(f_{1,0}(Vp(0))).
        %\label{p9_p11}
\end{equation}
Let $x=f_{1,0}(Vp(0))$ and we can get $f_2(x)=x$. Therefore,
\begin{equation}
	    f_{1,t'}(\sum_{t=0}^{\tau-1}Vp(t))=\sum_{t=0}^{\tau-1}f_{1,t}(Vp(t)),\forall t'.
        \label{p9_p11_new}
\end{equation}

\begin{lemma}
Eq. (\ref{p9_p11_new}) holds true for any value of $\{p(t),\tau,V,t'\}$ if and only if $f_{1,t}(x)=b_1(x+b_{0,t})$ with $b_1>0$.
\end{lemma}
\begin{proof}
To achieve the additive property of performance metrics across the time domain, the function $f$ must be linear. If $f$ is nonlinear, then the function value obtained through time averaging will not be equal to the function value evaluated at the average state (as indicated by Jensen's inequality).
\end{proof}

We do not differentiate whether the Lyapunov drift-plus-penalty function is scaled to $\mathcal Cb_1$ or $\mathcal C/b_1$. Consequently, for simplicity and without loss of generality, we set $b_1=1$ and have $f_{1,t}(x)=x+b_{0,t}$.
Based on the theoretical derivations presented above, the reward function suitable for RL, derived from the Lyapunov drift-plus-penalty method, is formulated as follows
\begin{equation}
	    \mathcal R(t)=-\frac{1}{2}\sum_{n=1}^N (Q^2_{n}(t+1)-Q^2_{n}(t))-Vp(t)-b_{0,t}.
\end{equation}

The task is simplified to determine $b_{0,t}$. Within a single time slot, $b_{0,t}$ is a constant and independent of any other quantities that may vary within the slot. Consequently, $b_{0,t}$ depends solely on fixed values within the time slot. In this study, we investigate arbitrary problems satisfying the given expression, where the only common fixed value across time slots is the queue state information, denoted as $Q(t)$. Therefore, we define $b_{0,t} = f_3(Q(t))$ as
\begin{equation}
	    \mathcal R(t)=-\frac{1}{2}\sum_{n=1}^N (Q^2_{n}(t+1)-Q^2_{n}(t))-Vp(t)-f_3(Q(t)).
        \label{reward_32}
\end{equation}
The objective is to optimize the long-term stability of the queue length. Hence, the specific time at which a particular queue length occurs is out of our concern. This leads to the following corollary.
\begin{lemma}%[Fibration]
If the queue lengths $Q_n(t)$ and $Q_n(t+1)$ are swapped while keeping the penalty term $p(t)$ unchanged, the reward in Eq.(\ref{reward_32}) should remain constant.
\label{lemma1}
\end{lemma}
\begin{proof}
The reward in Eq.(\ref{reward_32}) consists of queue stability part $-\frac{1}{2}\sum_{n=1}^N (Q^2_{n}(t+1)-Q^2_{n}(t))-f_3(Q(t))$ and penalty part $-Vp(t)$.

We swap the queue lengths $Q_n(t)$ and $Q_n(t+1)$ while keeping the penalty term $p(t)$ unchanged:
\begin{equation}
    \begin{aligned}
		&Q^*_n(t)=Q_n(t+1)\\
            &Q^*_n(t+1)=Q_n(t).
    \end{aligned}
\end{equation}

In a queueing system, swapping the queue lengths at two different times does not affect the queue stability. The reward in Eq.(\ref{reward_32}) should remain constant, as both the queue stability component and the penalty term remain constant.

\end{proof}
From Lemma \ref{lemma1}, we have $\mathcal R(t)=\mathcal R^*(t)$, where $R^*(t)$ is the reward when the queue lengths at the time slot $t$ and at the time slot $t+1$ are swapped:
\begin{equation}
    \begin{aligned}
		\mathcal R^*(t)=&-\frac{1}{2}\sum_{n=1}^N (Q^2_{n}(t)-Q^2_{n}(t+1))-Vp(t)\\
             &-f_3(Q(t+1)).
    \end{aligned}
\end{equation}
Therefore,

\begin{equation}
    \begin{aligned}
		&-\frac{1}{2}\sum_{n=1}^N (Q^2_{n}(t+1)-Q^2_{n}(t))-Vp(t)-f_3(Q(t))\\
		=&-\frac{1}{2}\sum_{n=1}^N (Q^2_{n}(t)-Q^2_{n}(t+1))-Vp(t)-f_3(Q(t+1)).
    \end{aligned}
\label{f3_1}
\end{equation}
After moving the terms related to $t$ to the right and $t+1$ to the left in Eq. (\ref{f3_1}) and simplifying, we can get
\begin{equation}
    f_3(Q(t+1))-\sum_{n=1}^NQ^2_n(t+1)=f_3(Q(t))-\sum_{n=1}^NQ^2_n(t).
\end{equation}
We can deduce that $f_3(Q(t))-\sum_{n=1}^NQ^2_n(t)$ is a constant, and we set $f_3(Q(t))-\sum_{n=1}^NQ^2_n(t)=b_0$. Therefore,
\begin{equation}
    f_3(Q(t))=\sum_{n=1}^NQ^2_n(t)+b_0.
\end{equation}

\subsection{Proposed LDPTRLQ method}

The constant term in the reward function for RL is not significant. Therefore, setting $b_0 = 0$ simplifies the expression without losing any generality. Hence,
\begin{equation}
    \begin{aligned}
		\mathcal R(t)=&-\frac{1}{2}\sum_{n=1}^N (Q^2_{n}(t+1)-Q^2_{n}(t))-Vp(t)-f_3(Q(t))\\
		=&-\frac{1}{2}\sum_{n=1}^N (Q^2_{n}(t+1)-Q^2_{n}(t))-Vp(t)\\
        &-\sum_{n=1}^NQ^2_n(t)-b_0\\
        =&-\frac{1}{2}\sum_{n=1}^N (Q^2_{n}(t+1)+Q^2_{n}(t))-Vp(t).
    \end{aligned}
    \label{R_final}
\end{equation}
The reward function in Eq. (\ref{R_final}) is the only reward function in Lemma \ref{lemma2} that satisfies a series of properties of RL in dealing with queueing problems. According to Theorem \ref{theorem3}, there exists at least one form of Eq. (\ref{ALL_R}) that is an optimal and static reward function. Therefore, the reward function in Eq. (\ref{R_final}) applies to all cases.

This form differs significantly from the objective functions in the widely used $\mathcal P3$ or $\mathcal P4$ algorithms. Therefore, we demonstrate that directly using the objective functions from $\mathcal P3$ or $\mathcal P4$ in RL will result in performance degradation. Furthermore, in the context of RL, the action is defined as $\alpha\in\mathcal{A}$, while the state $s\in\mathcal{S}$ encompasses system information, including the queue state.

Additionally, when the proposed RL algorithm is reduced to performing greedy optimization within a single time slot, we have
\begin{equation}
    \begin{aligned}
		&\mathcal P12:\min_{\alpha(t)}-\mathcal R(t) \\
		&\text {s.t.}\ \alpha(t)\in\mathcal A(t).
    \end{aligned}
\end{equation}

$\mathcal P12$ is equivalent to the original Lyapunov drift-plus-penalty optimization problem $\mathcal P3$, because $Q_{n}(t)$ are constant terms at time slot $t$. This equivalence validates the rationality of the proposed approach.

\subsection{Adaptive Lyapunov Weight Algorithm}
Although the Lyapunov weight $V$ should typically be predetermined in most problems, predetermining it is impractical in certain scenarios, while exhaustively scanning different values of $V$ is computationally prohibitive. Two popular approaches have been proposed to address similar issues. The first is an adaptive $V$ based on queue length: $V(t) = c \times Q^\beta(t)$, where $c$ is a pre-set coefficient, and $Q^\beta(t)$ is a positive correlation function with respect to queue length. The second is a water-filling-inspired approach: $V(t) = \max\{V_{\min}, \min\{V_{\max}, \kappa / ( \text{target-delay} - \text{current-delay})\}$, where $\kappa$ is a pre-set water-filling coefficient.

However, these methods are not applicable to the problem at hand, as RL requires $V$ to remain quasi-static. Moreover, tuning certain parameters in these approaches remains challenging. Motivated by the essence of the problem, we propose the following algorithm.

\begin{figure}[t]
\centering
\includegraphics[width=0.45\textwidth]{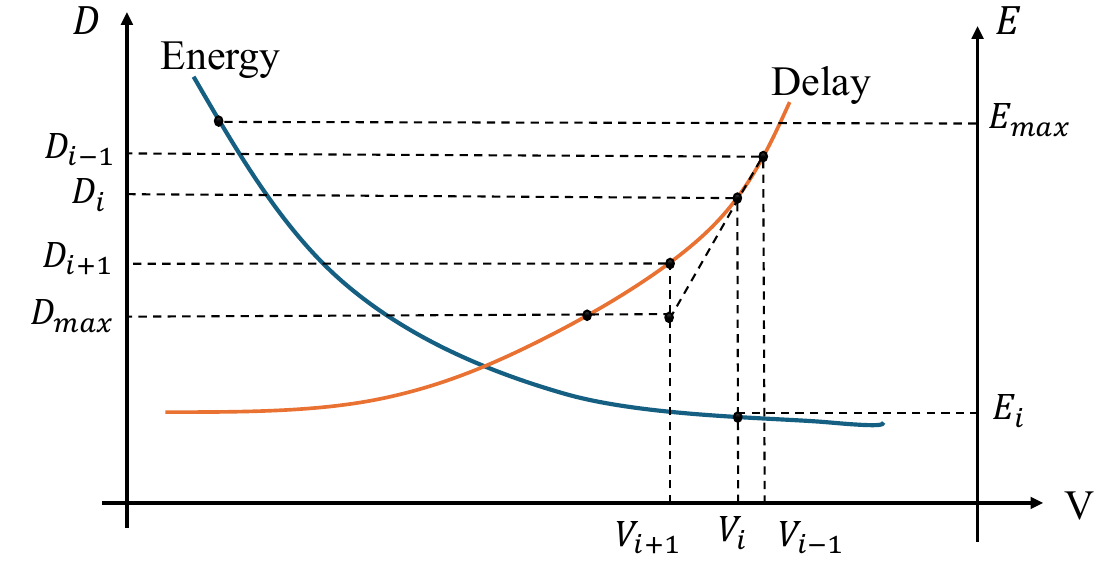}
\caption{Adaptive Lyapunov Weight Algorithm.}
\label{Adaptive_V}
\end{figure}

We assume that the overall objective is to satisfy the maximum delay constraint $D_{\max}$ and the maximum energy consumption constraint $E_{\max}$. The delay $D$ is related to the queue length $Q$, while the energy consumption $E$ is associated with the Lyapunov penalty term $p$. Specifically, the delay $D$ increases with larger values of $V$, whereas the energy consumption $E$ decreases as $V$ increases. To find an appropriate $V$ that balances these trade-offs, we employ the secant method illustrated in Fig. \ref{Adaptive_V} and Algorithm \ref{Adaptive_V_algo}.

\begin{algorithm}[t]
	\caption{Adaptive Lyapunov Weight Method}
        \renewcommand{\algorithmicrequire}{\textbf{Result:}}
	\renewcommand{\algorithmicensure}{\textbf{Input:}}
	\begin{algorithmic}[1]
        \ENSURE $D_{\max}$, $E_{\max}$.
		\REQUIRE $V$.
		\STATE Initialize a random Lyapunov weight $V_0>0$, the scaling factor $\eta_a=0.1$, the maximum exploration factor $\eta_{max}=0.5$, the minimum exploration factor $\eta_{min}=0.01$, and $i=0$.
            \STATE RL is performed using $V_0$, and $\{D_0,E_0\}$ are obtained.
            \WHILE{\{$D_0>D_{\max}$ and $E_0>E_{\max}$\}}
            \STATE $D_{\max}=(1+\eta_a)D_{\max}$. $E_{\max}=(1+\eta_a)E_{\max}$.
            \ENDWHILE
            \STATE If $D_0>D_{\max}$ then $V_1=(1-\eta_a)V_0$.
            \STATE If $E_0>E_{\max}$ then $V_1=(1+\eta_a)V_0$.
		\REPEAT
            \STATE $i=i+1$.
            \STATE RL is performed using $V_i$, and $\{D_i,E_i\}$ are obtained.
            \WHILE{\{$D_i> D_{\max}$ and $E_i> E_{\max}$\}}
            \STATE $D_{\max}=(1+\eta_a)D_{\max}$. $E_{\max}=(1+\eta_a)E_{\max}$.
            \ENDWHILE
		  \STATE If $D_i>D_{\max}$ then $V_{i+1}=\text{clip}(V_i+\frac{D_{\max}-D_i}{D_i-D_{i-1}}(V_i-V_{i-1}),(1-\eta_{max})V_i,(1-\eta_{min})V_i)$.
            \STATE If $E_i>E_{\max}$ then $V_{i+1}=\text{clip}(V_i+\frac{E_{\max}-E_i}{E_i-E_{i-1}}(V_i-V_{i-1}),(1+\eta_{min})V_i,(1+\eta_{max})V_i)$.
		\UNTIL \{$D_i\leq D_{\max}$ and $E_i\leq E_{\max}$\} or upper limit of iterations.
	\end{algorithmic}
	\label{Adaptive_V_algo}
\end{algorithm}

To ensure robust convergence and prevent system instability during the nascent exploration phase, we implement a structured initialization and threshold relaxation strategy. Since the secant method requires two distinct initial points to estimate the local gradient, the algorithm begins by stochastically generating an initial weight $V_0$. Subsequently, a second point $V_1$ is derived by applying a fixed exploration step size to $V_0$.

If the initial system performance $\{D_i, E_i\}$ violates both the delay constraint $D_{\max}$ and the energy constraint $E_{\max}$ simultaneously, an adaptive scaling mechanism is triggered to relax these thresholds by a factor $\eta_a$ until at least one constraint becomes feasible, thereby establishing a valid starting region for the iterative process.

In scenarios where the delay constraint $D_{\max}$ is violated, the update rule for $V_{i+1}$ is formulated based on the geometric principles of the Secant method. By assuming that the points $(D_{i-1}, V_{i-1})$, $(D_i, V_i)$, and $(D_{\max}, V_{i+1})$ are collinear as illustrated in Fig. \ref{Adaptive_V}, the next iteration value is derived through linear interpolation as follows:
\begin{equation}
\begin{aligned}
    V_{i+1}=&\text{clip}(V_i+\frac{D_{\max}-D_i}{D_i-D_{i-1}}(V_i-V_{i-1}),\\
    &(1-\eta_{\max})V_i,(1-\eta_{\min})V_i),
\end{aligned}
\end{equation}
where the function $\text{clip}(x, a, b) = \max(a, \min(x, b))$ restricts the variable to a predefined interval $[a, b]$.

The introduction of the $\text{clip}$ function is critical. It prevents excessive fluctuations in $V_i$ that could lead to algorithmic divergence. Furthermore, by enforcing a minimum exploration bound $(1-\eta_{\min})V_i$, we ensure that the algorithm does not prematurely stagnate, thereby driving the delay $D_i$ to converge toward or below the target $D_{\max}$ rather than merely approaching it asymptotically.

A symmetrical update logic is applied when the energy consumption constraint $E_{\max}$ is violated.

\subsection{Saturated Queueing Algorithm}
\begin{figure}[t]
\centering
\includegraphics[width=0.45\textwidth]{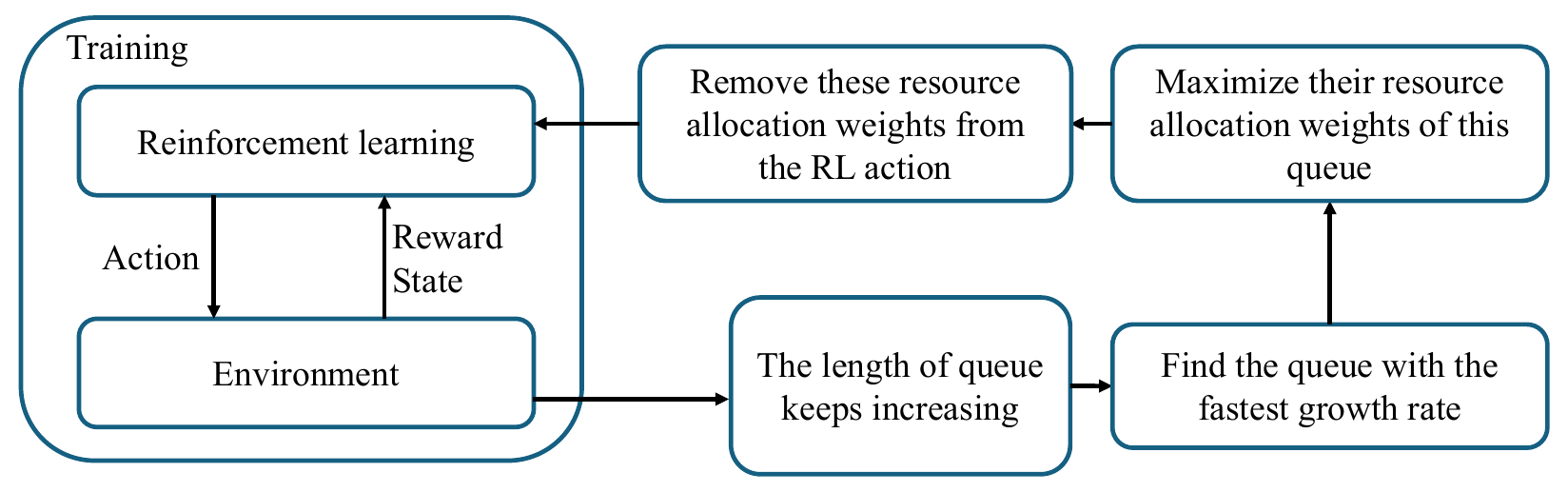}
\caption{Saturated Queueing Algorithm.}
\label{Saturated_Queuing}
\end{figure}

RL training requires an effective reward function. However, when the arrival rate approaches the service rate, the queue length can easily grow unbounded. Under such conditions, unbounded queue growth renders the reward function partially ineffective, as the penalties or rewards lose their discriminative power. Consequently, in scenarios where certain queues operate near saturation, the RL policy tends to oscillate among invalid solutions, significantly hindering algorithm convergence. To address this challenge when the arrival rate approaches the service capacity limit, we design a dedicated bailout mechanism as shown in Fig. \ref{Saturated_Queuing}.

This algorithm is triggered when the RL training fails to converge due to unbounded growth in certain queues. It first identifies the queue exhibiting the fastest growth rate, then sets the resource allocation coefficient for that queue to its maximum value, and subsequently removes this coefficient from the action space of the RL agent. The objective of this mechanism is to detect saturated queues and directly maximize their resource allocation, thereby preventing unbounded queue growth and stabilizing the system.

\section{Numerical Results and Analysis}
We introduce two queueing-related practical problems to evaluate the performance of our proposed algorithm. 
\subsection{Mobile Edge Computing}

We consider a system model for mobile edge computing, comprising $K$ mobile users and a single base station as shown in Fig. \ref{LDPP_f1}. The base station is equipped with an edge computing server dedicated to processing computational tasks. Each mobile user generates computational tasks in each time slot with a certain probability, where task arrivals follow a Poisson process with an arrival rate of $\lambda$. The size of each task is uniformly distributed over the interval $(0, d_{\text{max}})$. For each mobile user, a task queue $Q_k(t)$ is maintained, with a local computing speed (bits/s) denoted by $c^L_k(t)$ and an upper bound on the computing speed (bits/s) given by $c^L_{\text{max},k}(t)$. 

\begin{figure}[t]
\centering
\includegraphics[width=0.45\textwidth]{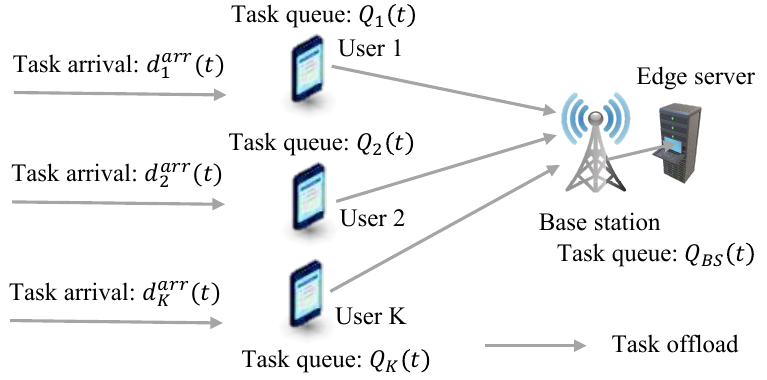}
\caption{System model of mobile edge computing.}
\label{LDPP_f1}
\end{figure}
The evolution of the queue lengths at the $k^{th}$ mobile user at time slot $t$ can be described by the following queueing model:
\begin{equation}
	Q_k(t+1)=\max\{Q_k(t)-d^{\text{dep}}_k(t),0\}+d^{\text{arr}}_k(t), \forall k,
\end{equation}
where $d^{\text{arr}}_k(t)$ is the total length of the newly generated task demand at time slot $t$, and $d^{\text{dep}}_k(t)$ represents the number of tasks departing from the task queue at time slot $t$, which comprises the tasks processed locally within the time slot $t$ and those offloaded to the edge computing server:
\begin{equation}
	d^{\text{dep}}_k(t)=R_{k}(t)\Delta T +c^L_k(t)\Delta T, \forall k,
\end{equation}
where $\Delta T$ is the length of each time slot, and $R_{k}(t)$ is the offloading data rate (bit/s). Task offloading is facilitated through wireless communication. We adopt a fundamental wireless communication model wherein each user is assigned an independent orthogonal channel, each with a bandwidth of $B$. The transmission rate from user $k$ to the base station, which corresponds to the task offloading rate, is given by
\begin{equation}
	R_k(t) = B \log_2 \left( 1 + \frac{\omega_k^{\text{WL}}(t)P_k(t)}{\sigma} \right), \forall k,
\end{equation}
where $\omega_k^{\text{WL}}(t)$ denotes the channel gain at time slot $t$, $P_k(t)$ represents the transmission power from user $k$ to the base station at time slot $t$, and $\sigma$ is the channel noise power. The evolution of the task queue lengths at the edge server at the base station at time slot $t$ can be described by the following queueing model
\begin{equation}
	Q_\text{BS}(t+1)=\max\{Q_\text{BS}(t)-d^{\text{dep}}_\text{BS}(t),0\}+d^{\text{arr}}_\text{BS}(t),
\end{equation}
where $d^{\text{arr}}_\text{BS}(t)$ is the total length of the newly received task demand at the base station at time slot $t$: $d^{\text{arr}}_\text{BS}(t)=\sum_{k=1}^KR_{k}(t)\Delta T$,
and $d^{\text{dep}}_\text{BS}(t)$ represents the number of tasks departing from the task queue $Q_\text{BS}(t)$ at time slot $t$, which comprises the tasks processed within the time slot $t$ at the edge server: $d^{\text{dep}}_\text{BS}(t)=c^E_\text{BS}(t)\Delta T$. $c^E_\text{BS}(t)$ is the task processing speed (bit/s) at the edge server at the base station at time slot $t$ with an upper bound given by $c^E_{\text{max},\text{BS}}(t)$. The energy consumption within each time slot comprises both the energy expended on communication and the energy expended on computation:
\begin{equation}
    \begin{aligned}
		E^{\mathbb {A}}(t)=&E^{\mathbb WC}(t)+E^{\mathbb {CC}}(t)\\
		=&\sum_{k=1}^KP_k(t)\Delta T+\sum_{k=1}^K\eta_kc^L_k(t)\Delta T+\eta_\text{BS}c^E_\text{BS}(t)\Delta T,
    \end{aligned}
\end{equation}
where $\eta_k$ is the computing energy consumption per bit when the task is processed at the $k^{th}$ mobile user, and $\eta_\text{BS}$ is the computing energy consumption per bit when the task is processed at the edge server at the base station.

The reward function at time slot $t$ of the proposed LDPTRLQ algorithm, as derived in Eq. (\ref{R_final}) of Section III, is applied to this specific edge computing problem, resulting in the following formula
\begin{equation}
    \begin{aligned}
		\mathcal R(t)=&-\frac{1}{2}\sum_{k=1}^N (Q^2_k(t+1)+Q^2_k(t))\\
        &-\frac{1}{2}(Q^2_\text{BS}(t+1)+Q^2_\text{BS}(t))-VE^{\mathbb {A}}(t).
    \end{aligned}
\end{equation}
The action at time slot $t$ contains the local computing speed of the mobile user, the computing speed of the edge computing server, and the power allocation for wireless communication:
\begin{equation}
	\boldsymbol a(t)=\{c^L_k(t), c^E_\text{BS}(t), P_k(t),\forall k\}.
\end{equation}
To ensure the physical feasibility of decisions and maintain training stability, we incorporate a projection layer into the output of the actor network. In the considered MEC scenario, an action is deemed invalid if the sum of the offloaded task volume and the locally processed volume exceeds the total available workload in the task buffer. The action actually used is the closest legal action. The state at time slot $t$ comprises the task queue length at the mobile user, the task queue length at the edge computing server, and the wireless channel gain:
\begin{equation}
	\boldsymbol s(t)=\{Q_k(t), Q_\text{BS}(t), \omega_k^{\text{WL}}(t),\forall k\}.
\end{equation}
Our simulation system model consists of 1 base station with 1 edge computing server, serving 10 mobile users. The type of RL agent is proximal policy optimization (PPO). The PPO agent has 5 fully connected layers with 64 neurons per layer equipped with the ReLU activation function. We set the PPO clipping parameter to 0.2. The discount factor is 0.95. The maximum number of steps per episode is 500, and the maximum number of episodes is 1000. The simulation environment is equipped with an Intel® Core™ i9-13900HX CPU and an RTX 4080 GPU. The simulation settings are also shown in TABLE \ref{Simulation_settings}.

\begin{table}[t]
\renewcommand{\arraystretch}{1.3}
\caption{Simulation settings}
\label{Simulation_settings}
\centering
\begin{tabular}{cccc}
\hline
\bfseries Parameter & \bfseries Value & \bfseries Parameter & \bfseries Value\\
\hline
Bandwidth & 10kHz &  Noise & $3.16\times 10^{-11}$\\
Arrival model & Poisson &  Channel Model & Rayleigh\\
$c^L_{\text{max},k}$ & 1000 &  $c^E_{\text{max},\text{BS}}$ & 5000\\
$\lambda$ & 2 &  $K$ & 10\\
Activ. func. & ReLU &  PPO agent & 5 layers\\
Layer & 64 neurons & Mini batch & 128\\
Reply memory & 10000 &  Discount factor & 0.95\\
Steps per episode & 500 &  Maximum episode & 1000\\
\hline
\end{tabular}
\end{table}

To verify the efficiency of our proposed algorithm, we performed a comparative analysis against several different algorithms. These comparisons aim to highlight the performance enhancements achieved through our method under various standard scenarios used in the field: 

\begin{itemize}
\item Original LDPRLQ: This is the original Lyapunov Drift Plus Penalty with RL (LDPRLQ) algorithm, as utilized in several academic papers \cite{dai2020deep}. It directly uses the negative of the Lyapunov drift plus penalty function as the reward for RL.

\item Simplified LDPRLQ: A variant adopted in \cite{bi2021lyapunov}, where the Lyapunov drift plus penalty function is approximated as follows
\begin{equation}
    \mathcal C(t)=\sum_{n=1}^NQ_{n}(t)(Q_{n}(t+1)-Q_{n}(t)) +Vp(t).
\end{equation}
The negative of $\mathcal C(t)$ is used as the reward for RL. 

\item LERL: Latency-energy tradeoff with RL. An algorithm unrelated to the Lyapunov drift plus penalty framework, which optimizes a weighted sum of latency and energy consumption, as presented in \cite{xu2023energy}. In the context of our problem, latency corresponds to the queue length, thus directly optimizing the weighted sum of queue length and the penalty term.

\item PPO-Lagrangian: Proximal Policy Optimization Lagrangian is a widely recognized baseline in constrained RL. It extends the standard PPO algorithm to handle CMDP by incorporating the Lagrangian multiplier method. The objective function is augmented with a penalty term, where the Lagrange multiplie scales the cost surrogate based on the degree of constraint violation.

\item LDPTRLQ(DQN): A modification of our proposed algorithm where the default PPO agent is replaced with a Deep Q-Network (DQN) agent.

\end{itemize}

\subsubsection{Convergence speed comparison}
\begin{figure}[t]
\centering
\includegraphics[width=0.45\textwidth]{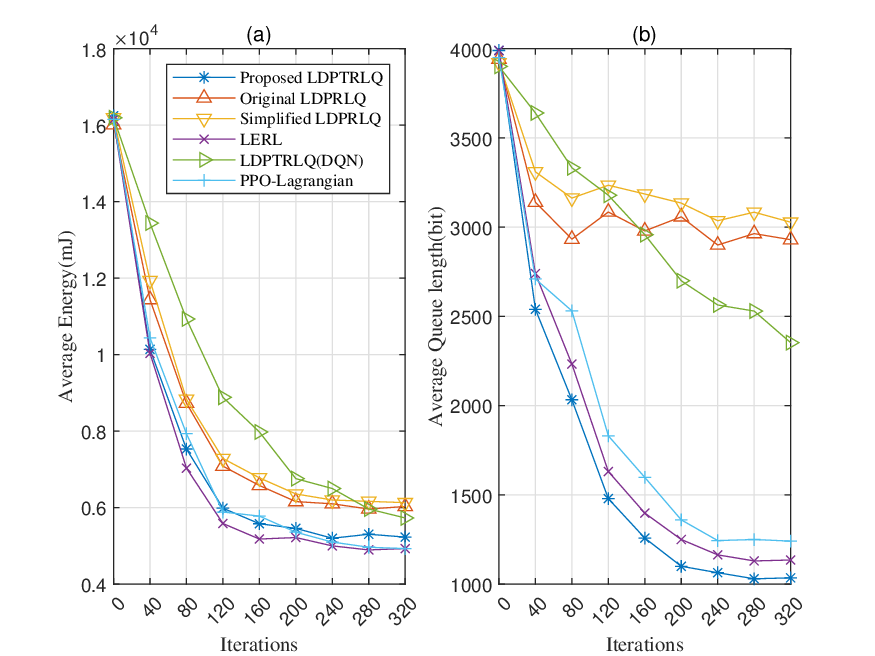}
\caption{The convergence behavior of our proposed LDPTRLQ with benchmarks.}
\label{mec_converge}
\end{figure}
Fig. \ref{mec_converge} illustrates the convergence behavior of the proposed LDPTRLQ algorithm alongside four baseline methods. Since the trade-off coefficient in the LERL algorithm differs in meaning from the Lyapunov weight $V$ in our proposed approach, a preliminary traversal of LERL's trade-off coefficient was conducted. This enabled the selection of a coefficient value that aligns the energy consumption and queue length metrics with those of our algorithm, facilitating a fair comparison under consistent conditions. Simulation results reveal that the proposed LDPTRLQ algorithm achieves a balanced trade-off, significantly reducing both average energy consumption and average queue length. Its convergence speed surpasses that of Original LDPRLQ, Simplified LDPRLQ, and LDPTRLQ(DQN), demonstrating enhanced optimization efficiency. While the LERL algorithm yields results comparable to our method, it requires careful tuning of the trade-off coefficient to ensure queue stability, whereas our algorithm inherently maintains stability without such adjustments. Although PPO-Lagrangian exhibits competitive performance comparable to LERL and the proposed LDPRLQ in terms of energy and latency, its efficacy heavily relies on the manual pre-specification of constraint violation penalties for different scenarios. Furthermore, the slower convergence of LDPTRLQ(DQN) compared to the proposed LDPTRLQ suggests that, under these problem conditions, the PPO agent outperforms the DQN agent. The superior convergence speed of our LDPTRLQ over Original LDPRLQ and Simplified LDPRLQ corroborates our theoretical derivations, underscoring deficiencies in these baseline methods that our approach effectively mitigates.

\subsubsection{Impact of different Lyapunov weight of proposed LDPTRLQ and benchmarks}
\begin{figure}[t]
\centering
\includegraphics[width=0.45\textwidth]{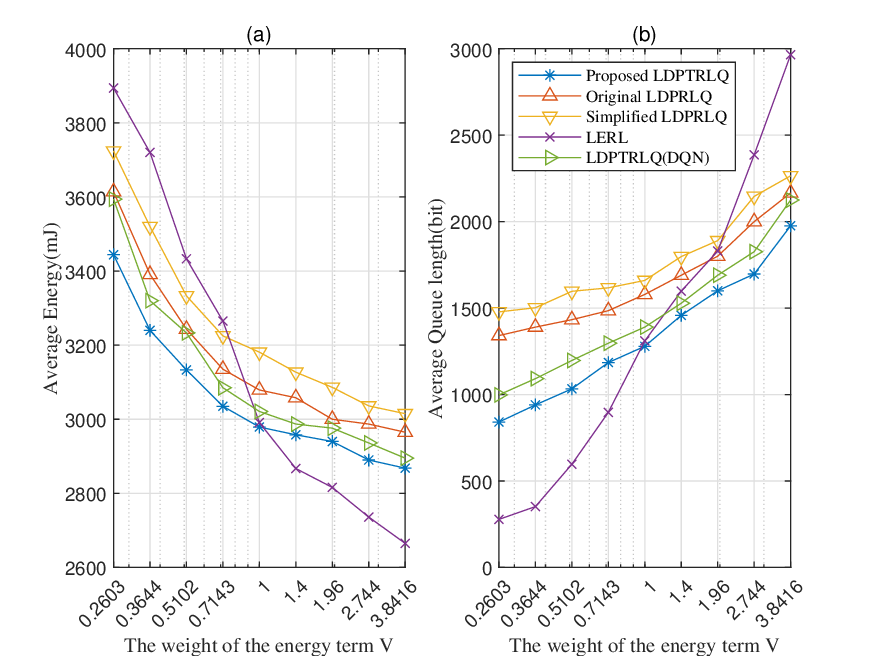}
\caption{The energy and queue length behavior of our proposed LDPTRLQ with benchmarks under varying weights.}
\label{MEC_V}
\end{figure}
Fig. \ref{MEC_V} presents the performance of the proposed LDPTRLQ algorithm alongside four baseline methods under varying Lyapunov weight conditions. Given that the trade-off weight in the LERL algorithm differs in interpretation from the Lyapunov weight in our approach, a standardization of the weight factors was performed in advance. Specifically, a coarse traversal of the trade-off coefficient in LERL was conducted, and the weight yielding energy consumption and queue length values closest to those of the proposed LDPTRLQ algorithm was normalized to 1. This normalization enables a consistent comparison across the algorithms. Simulation results demonstrate that the proposed LDPTRLQ algorithm outperforms Original LDPRLQ and Simplified LDPRLQ in terms of both average queue length and average energy consumption across all tested conditions. This improvement validates our theoretical findings, confirming that the proposed algorithm enhances the Lyapunov Drift-Plus-Penalty framework for RL applications. In contrast, LDPTRLQ(DQN) consistently exhibits inferior performance relative to the proposed LDPTRLQ, indicating that the PPO agent is more effective than the DQN agent in this application context. While the LERL algorithm achieves performance comparable to that of LDPTRLQ, it exhibits greater sensitivity to weight variations, making it challenging to identify a weight that ensures queue stability.

\subsubsection{Impact of different number of users of proposed LDPTRLQ and benchmarks}
\begin{figure}[t]
\centering
\includegraphics[width=0.45\textwidth]{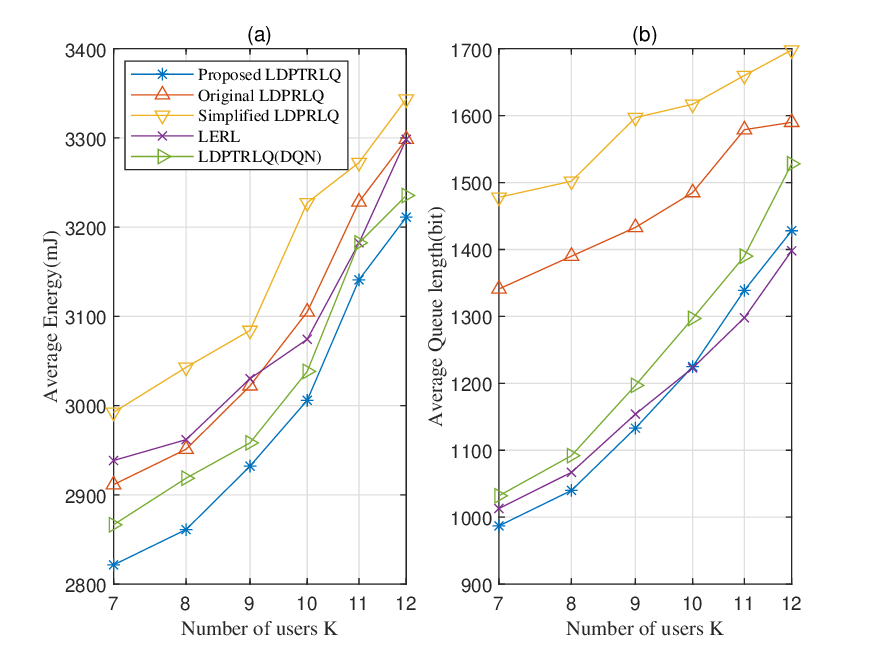}
\caption{The energy and queue length behavior of our proposed LDPTRLQ with benchmarks under varying number of users.}
\label{MEC_user}
\end{figure}
Fig. \ref{MEC_user} presents the performance of the proposed LDPTRLQ algorithm alongside four baseline methods under varying number of users. The simulation results demonstrate that the proposed LDPTRLQ algorithm consistently outperforms the original LDPRLQ and the simplified LDPRLQ variants across all test configurations in terms of both average queue length and average energy consumption. This superiority highlights the efficacy of integrating trust region constraints with the Lyapunov-based framework. Specifically, the performance of LDPTRLQ(DQN) is observed to be systematically inferior to the proposed PPO-based LDPTRLQ. While the LERL algorithm achieves competitive results in certain scenarios, its performance is marginally lower than that of LDPRLQ. Furthermore, as the number of users increases, the average energy consumption and queue length for all algorithms exhibit an upward trend. However, the growth rate remains within a reasonable and manageable range, which validates the scalability of the proposed framework in handling high-dimensional state and action spaces.

\subsubsection{The queue length standard deviation of proposed LDPTRLQ and benchmarks}
\begin{figure}[t]
\centering
\includegraphics[width=0.45\textwidth]{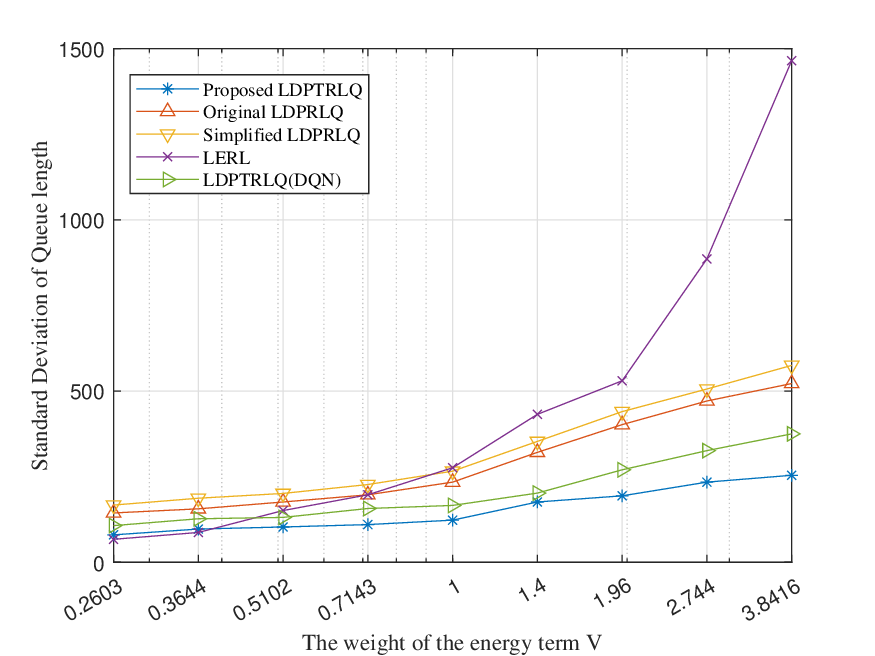}
\caption{The standard deviation queue length behavior of our proposed LDPTRLQ with benchmarks under varying weights.}
\label{MEC_sigma}
\end{figure}

Fig. \ref{MEC_sigma} illustrates the queue length standard deviation of the proposed LDPTRLQ algorithm and four baseline methods under varying Lyapunov weight conditions. The same weight factor standardization approach as described previously was applied. The results indicate that the proposed LDPTRLQ algorithm achieves the smallest queue length standard deviation in most scenarios, with LERL exhibiting a marginally lower standard deviation only when the queue weight factor is significantly elevated. This suggests that our proposed algorithm ensures superior queue stability compared to all baseline methods. The enhanced stability of LDPTRLQ stems from its incorporation of the queue length as a squared term in the RL reward function, in contrast to LERL, which uses a linear term. Consequently, LERL is less sensitive to occasional excessive queue lengths, functioning primarily as a trade-off mechanism between queue length and energy consumption rather than an optimization strategy prioritizing queue stability. Similarly, the proposed LDPTRLQ algorithm consistently outperforms Original LDPRLQ and Simplified LDPRLQ in terms of queue length standard deviation, demonstrating an improvement over conventional approaches that integrate Lyapunov Drift-Plus-Penalty with RL. Furthermore, the superior performance of LDPTRLQ over LDPTRLQ(DQN) reinforces the finding that, in this application context, the PPO agent outperforms the DQN agent.

\subsubsection{Convergence behavior of the adaptive Lyapunov weight algorithm}
\begin{figure}[t]
\centering
\includegraphics[width=0.45\textwidth]{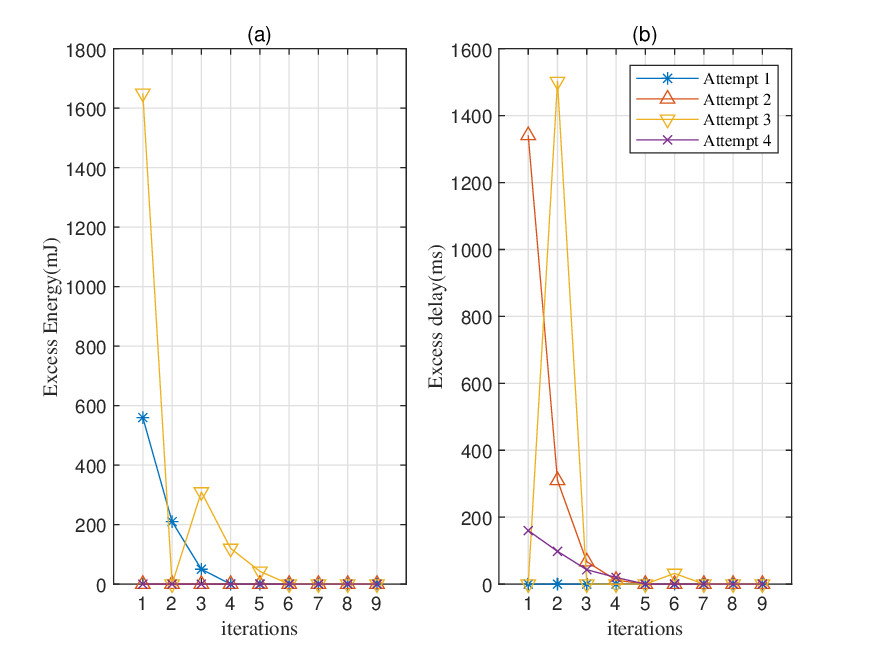}
\caption{The convergence behavior of our proposed adaptive Lyapunov weight algorithm.}
\label{MEC_add_V}
\end{figure}

Fig. \ref{MEC_add_V} illustrates the convergence performance of our proposed dynamic Lyapunov weight algorithm across four independent runs. The convergence speed is influenced by the initial random value of the Lyapunov weight $V$. In all cases, the algorithm rapidly converges to a solution where neither the energy consumption exceeds the prescribed maximum $E_{\max}$ nor the delay surpasses the maximum $D_{\max}$. This demonstrates the effectiveness of the proposed dynamic Lyapunov weight adjustment algorithm in efficiently identifying an appropriate weight $V$.

\subsubsection{Convergence behavior of the saturated queueing algorithm}
\begin{figure}[t]
\centering
\includegraphics[width=0.45\textwidth]{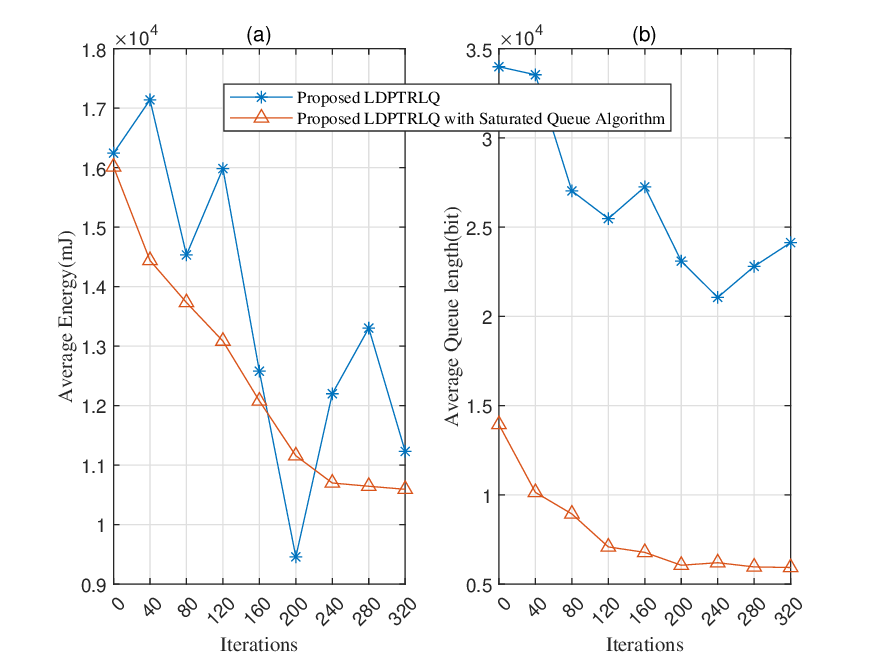}
\caption{The convergence behavior of our proposed saturated queueing algorithm.}
\label{MEC_sat_queue}
\end{figure}

Fig. \ref{MEC_sat_queue} illustrates the convergence performance of the proposed saturated queue bailout mechanism. In the simulation, one user is configured with a high task arrival rate approaching its computational capacity limit. The results show that, without the saturated queue mechanism, the RL algorithm experiences considerable difficulty in converging. When the policy explores actions that lead to unbounded queue accumulation, the reward function partially loses its discriminative power, thereby hindering effective policy improvement.
In contrast, with the proposed saturated queue bailout mechanism enabled, the algorithm promptly identifies the saturated user and directly sets its resource allocation coefficient to the maximum value. This intervention effectively prevents unbounded queue growth and enables the overall algorithm to achieve stable and proper convergence.

\subsection{Network Routing}

\begin{figure}[t]
\centering
\includegraphics[width=0.45\textwidth]{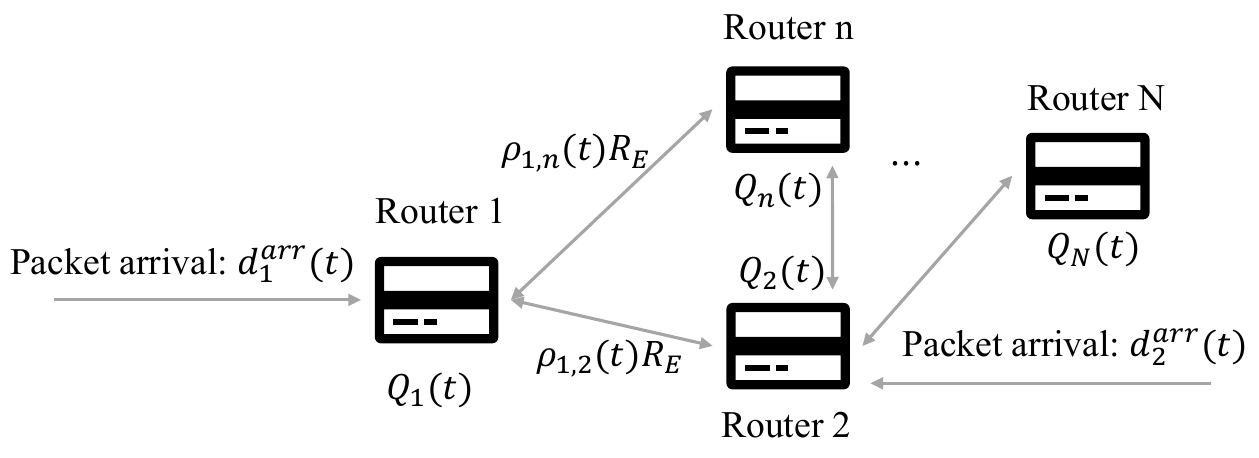}
\caption{System model of network routing.}
\label{topologies}
\end{figure}

In our simulation, we developed a method for constructing a randomly generated network topology for analysis as shown in Fig. \ref{topologies}. The generation process begins with the creation of $20$ nodes, each assigned a random two-dimensional coordinate $(x, y)$. The values for $x$ and $y$ are uniformly distributed between $0$ and $100$, ensuring that the node distribution covers the entire designated area evenly. Each node is programmed to potentially connect to its $5$ nearest nodes. The probability of each of these connections being established is set at $50\%$, introducing a stochastic element to the network structure that mimics real-world irregularities and network dynamics. If the topology graphs are separated, we connect the closest nodes between them. The achievable data rate with each connection, denoted as $R$, is calculated using the formula $R=\rho R_E$, where $\rho\in[0.3,1.7]$ represents a scaling factor and $R_E$ is the average data rate between the nodes. The uniformly distributed scaling factor $\rho$ allows the connection speed to reflect a range of potential network conditions, from favorable to adverse.

In \cite{neely2010stochastic}, the Lyapunov drift theory (with $V = 0$, i.e., without any penalty term) was employed to mathematically analyze the conventional backpressure routing algorithm. These studies demonstrated that the backpressure routing algorithm is consistent with the Lyapunov drift framework. Specifically, under the condition $V = 0$ , we can rewrite $\mathcal P4$ as
\begin{equation}
    \begin{aligned}
		&\mathcal P12:\min_{\alpha(t)}\sum_{n=1}^N Q_{n}(t)(d_n^{\text{arr}}(t)-d_n^{\text{dep}}(t)) \\
		&\text {s.t.}\ \alpha(t)\in\mathcal A(t),
    \end{aligned}
\end{equation}
where $\alpha(t)$ denotes the routing action. These patterns are designed such that each queue has exactly one next-hop routing target. Problem $\mathcal P12$ is formulated as a linear programming problem, which can be solved by the CVX optimization toolbox. This solution approach is equivalent to the backpressure routing algorithm. Similarly, our proposed LDPTRLQ algorithm is capable of addressing this routing problem.

\subsubsection{Impact of different arriving rate of proposed LDPTRLQ and benchmarks}
\begin{figure}[t]
\centering
\includegraphics[width=0.45\textwidth]{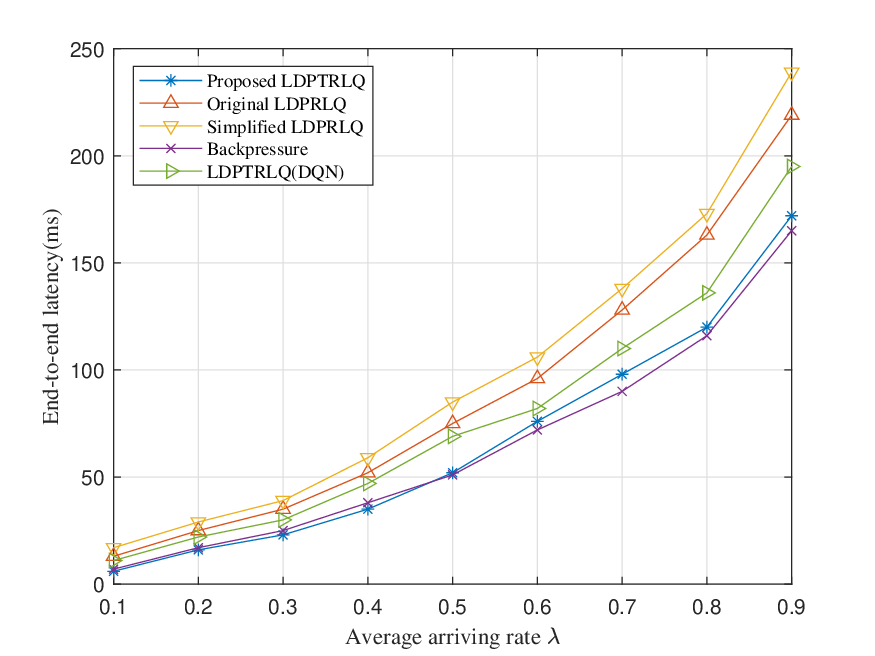}
\caption{The end-to-end latency versus the average  arriving rate of our proposed LDPTRLQ and benchmarks.}
\label{routing_lambda}
\end{figure}
The performance of our proposed LDPTRLQ algorithm, as illustrated in Fig. \ref{routing_lambda}, closely approximates that of the conventional backpressure routing algorithm. Notably, under low network congestion, LDPTRLQ exhibits a slight performance advantage over the backpressure algorithm. However, despite this advantage under specific conditions, the computational resource consumption of our proposed algorithm significantly exceeds that of the backpressure algorithm, which requires minimal computational resources. Consequently, in the absence of a penalty term, the LDPTRLQ algorithm does not outperform the backpressure algorithm because the problem reduces to a pure queue-stability task for which the backpressure algorithm is already known to provide a near-optimal solution with low computational overhead. Furthermore, the higher latency observed with LDPTRLQ(DQN) suggests that the PPO agent outperforms the DQN agent in this environment. The numerical results also demonstrate that our proposed LDPTRLQ algorithm exceeds those of other algorithms combining Lyapunov drift-plus-penalty with RL.

\subsubsection{Impact of different Lyapunov weight of proposed LDPTRLQ and benchmarks}
\begin{figure}[t]
\centering
\includegraphics[width=0.45\textwidth]{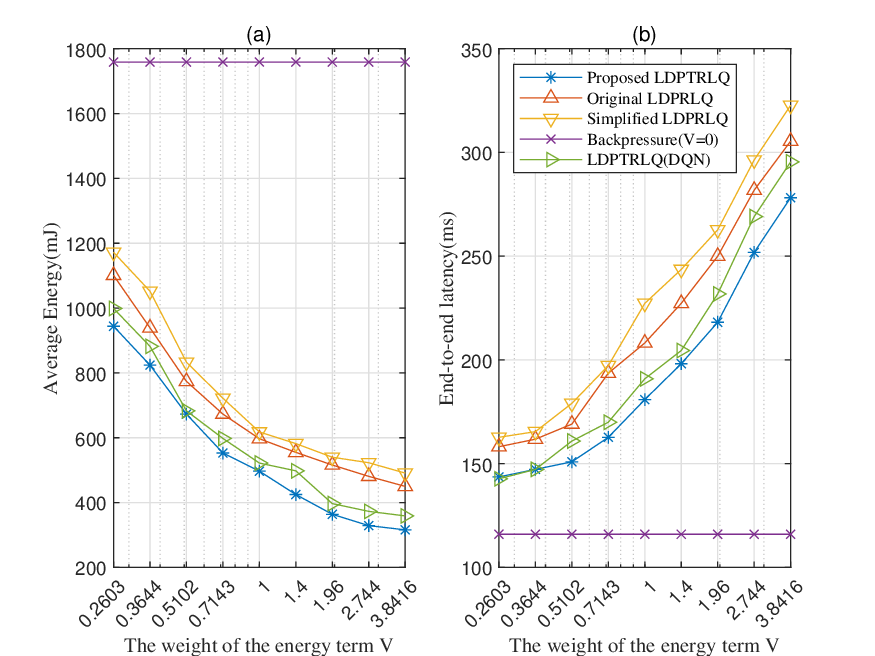}
\caption{The end-to-end latency versus the weight of energy term V our proposed LDPTRLQ and benchmarks.}
\label{routing_V}
\end{figure}
We introduce the energy model $E(t) = \eta_2 R^2(t) + \eta_1 R(t)$,
where $\eta_1$ and $\eta_2$ are known coefficients, and $R(t)$ is the transmission data rate at time slot $t$. Furthermore, we subsequently address the comprehensive Lyapunov drift-plus-penalty optimization framework. Incorporating complex penalty terms into the backpressure routing algorithm is challenging due to the non-convex nature of the optimization problem, which renders it difficult to solve. Consequently, for the backpressure routing algorithm, we fix the trade-off parameter $V = 0$. As illustrated in Fig. \ref{routing_V}, simulation results demonstrate that the proposed LDPTRLQ algorithm outperforms all other algorithms except the backpressure routing algorithm, highlighting its superiority over existing approaches combining Lyapunov drift-plus-penalty with RL. Moreover, while accounting for energy consumption, our algorithm achieves latency performance comparable to that of the backpressure algorithm to a certain extent. This ability to incorporate complex optimization factors while maintaining effective routing performance represents a key advantage of the proposed LDPTRLQ algorithm over the backpressure routing algorithm.

TABLE \ref{Simulation_comparison} summarizes the strengths of our proposed algorithm and various benchmarks.
\begin{table*}[t]
\renewcommand{\arraystretch}{1.3}
\caption{Algorithm comparison}
\label{Simulation_comparison}
\centering
\begin{tabular}{cccc}
\hline
\bfseries Algorithm & \bfseries Guarantee of Queue stability & \bfseries Trade-off between two objectives  & \bfseries At least one metric is optimal\\
\hline
Original LDPRLQ & \checkmark &  \checkmark & $\times$\\
Simplified LDPRLQ  & \checkmark & \checkmark & $\times$\\
LERL  & $\times$ &  \checkmark & \checkmark\\
PPO-Lagrangian & $\times$ &  \checkmark & \checkmark\\
LDPTRLQ(DQN)  & \checkmark & \checkmark & $\times$\\
Backpressure & \checkmark & $\times$  & \checkmark\\
Proposed LDPTRLQ & \checkmark & \checkmark & \checkmark\\
\hline
\end{tabular}
\end{table*}

\section{Conclusion}
In this paper, we proposed LDPTRLQ, an enhanced algorithm integrating the Lyapunov Drift-Plus-Penalty framework with RL to address the optimization challenges posed by emerging complicated systems. Through theoretical derivations, we established an effective approach that outperforms conventional methods by effectively combining the short-term greediness of Lyapunov Drift-Plus-Penalty with the long-term optimization capabilities of RL. Simulation results in the MEC environment demonstrate that LDPTRLQ consistently surpasses baseline algorithms. These findings validate our theoretical analysis, highlighting the algorithm's robustness and efficiency. While LERL exhibits competitive performance, its sensitivity to weight tuning underscores LDPTRLQ's practical advantage in maintaining queue stability without extensive parameter adjustments. Simulation results in the routing environment demonstrate that, compared to the conventional backpressure routing algorithm, the proposed LDPTRLQ algorithm effectively accommodates a wide range of complex optimization objectives. Furthermore, it outperforms other algorithms that integrate Lyapunov drift-plus-penalty with RL. Future work will focus on validating these results in real-world IoT deployments and exploring adaptive weight mechanisms to further enhance performance under dynamic conditions.

\ifCLASSOPTIONcaptionsoff
  \newpage
\fi



\bibliographystyle{IEEEtran}
\bibliography{IEEEabrv,bare_jrnl}



% \end{thebibliography}

%\begin{IEEEbiography}[{\includegraphics[width=1in,height=1.25in,clip,keepaspectratio]{mshell}}]{Wenhan Xu}



%\vfill


\end{document}